\documentclass[runningheads]{llncs}
\usepackage[T1]{fontenc}
\usepackage{graphicx}
\usepackage{hyperref}
\usepackage{color}

\usepackage{nth}
\usepackage{subcaption}
\usepackage[noend]{algorithm2e}
\RestyleAlgo{ruled}
\usepackage[most]{tcolorbox}

\usepackage{booktabs}
\usepackage{amssymb}

\begin{document}
\title{Investigating Low-Cost LLM Annotation for~Spoken Dialogue Understanding Datasets}
\titlerunning{LLM Spoken Dialogue Annotation}

\author{Lucas Druart\inst{1,2}, Valentin Vielzeuf\inst{2}\orcidID{0000-0002-2987-2510} \\ Yannick Estève\inst{1}\orcidID{0000-0002-3656-8883}}
\authorrunning{L. Druart et al.}

\institute{Laboratoire d'Informatique d'Avignon (LIA), Avignon, France \email{first.last@univ-avignon.fr} \and Orange Innovation, Rennes, France \\ \email{first.last@orange.com}}
\maketitle              % typeset the header of the contribution
\begin{abstract}
    In spoken Task-Oriented Dialogue (TOD) systems, the choice of the semantic representation describing the users' requests is key to a smooth interaction. Indeed, the system uses this representation to reason over a database and its domain knowledge to choose its next action. The dialogue course thus depends on the information provided by this semantic representation. While textual datasets provide fine-grained semantic representations, spoken dialogue datasets fall behind. This paper provides insights into automatic enhancement of spoken dialogue datasets' semantic representations. Our contributions are three fold: (1) assess the relevance of Large Language Model fine-tuning, (2) evaluate the knowledge captured by the produced annotations and (3) highlight semi-automatic annotation implications.

\end{abstract}

\section{Introduction}
    % Spoken Dialogue annotated data is necessary for applied models but often textual datasets
    Digitization enables many tasks to be automated, nevertheless users sometimes require assistance to perform complex tasks such as making a reservation at a restaurant or booking a hotel room. Task-Oriented Dialogue (TOD) systems are designed to assist such users. A common approach to implement them is to break the problem down to three iterative steps \cite{TurDeMori_SDS}: updating the system's understanding of the users' needs, reasoning over a database and domain knowledge to choose the next action and providing the user an answer. Those systems often rely on transfer learning which requires annotated datasets. However only few datasets provide aligned dialogue recordings with turn-level contextual semantic annotations. Therefore the dialogue understanding community has mainly focused on textual datasets creating a usage discrepancy \cite{faruqui-hakkani-tur-2022-revisiting}. 

    % Annotation schemes differences between written and spoken datasets
    Dialogue systems rely on a chosen semantic representation to infer the next action(s). The gap between textual semantic representations and spoken ones provides an explanation for the observed discrepancy. Indeed, spoken dialogue datasets often use a flat semantic representation embedded in the transcription, such as transcription span labels,  while textual ones propose more fine-grained structured representations such as Dialogue-Abstract Meaning Representation~\cite{bonial-etal-2020-dialogue} or Dialogue Meaning Representation \cite{hu-etal-2022-dialogue}. This paper proposes a method to automatically annotate dialogue datasets with fine-grained semantic representations.

    \begin{figure}[!htb]
            \begin{subfigure}[h]{\linewidth}
            \begin{flushleft}
                \textcolor{blue}{\texttt{<command-tache>}} j' aurais voulu réserver \textcolor{blue}{\texttt{>}} euh \textcolor{blue}{\texttt{<nombre-chambre-reservation>}} deux \textcolor{blue}{\texttt{> <chambre-type>}} chambres doubles \textcolor{blue}{\texttt{> <connectProp>}} et \textcolor{blue}{\texttt{> <nombre-chambre-reservation>}} une \textcolor{blue}{\texttt{> <chambre-type>}} chambre simple \textcolor{blue}{\texttt{> <sejour-nbNuit-reservation>}} pour les cinq jours \textcolor{blue}{\texttt{> <temps-jourFerie-reservation>}} de Noël à la Noël \textcolor{blue}{\texttt{>}} donc \textcolor{blue}{\texttt{<localisation-arrondissement-hotel>}} dans le huitième arrondissement \textcolor{blue}{\texttt{> <localisation-ville-hotel>}} de Paris \textcolor{blue}{\texttt{>}}
                \end{flushleft}
                \subcaption{Current labeled span annotation.}
            \end{subfigure}
            \begin{subfigure}[h]{\linewidth}
                \texttt{(\textcolor{red}{r1} / \textcolor{blue}{reservation} \\
                            \phantom{x}\hspace{1em}:\textcolor{green}{objet} (\textcolor{red}{h1} / \textcolor{blue}{hotel} \\
                            \phantom{x}\hspace{2em}:\textcolor{green}{chambre} (\textcolor{red}{e1} / \textcolor{blue}{et} \\
                            \phantom{x}\hspace{3em}:\textcolor{green}{arg1} (\textcolor{red}{c1} / \textcolor{blue}{chambre} \\
                            \phantom{x}\hspace{4em}:\textcolor{green}{type} "double"\\
                            \phantom{x}\hspace{4em}:\textcolor{green}{quantite} "deux") \\
                            \phantom{x}\hspace{3em}:\textcolor{green}{arg2} (\textcolor{red}{c2} / \textcolor{blue}{chambre} \\
                            \phantom{x}\hspace{4em}:\textcolor{green}{type} "simple"\\
                            \phantom{x}\hspace{4em}:\textcolor{green}{quantite} "une")) \\
                            \phantom{x}\hspace{2em}:\textcolor{green}{date-sejour} (\textcolor{red}{e2} / \textcolor{blue}{evenement} \\
                            \phantom{x}\hspace{3em}:\textcolor{green}{nom} "Noël") \\
                            \phantom{x}\hspace{2em}:\textcolor{green}{duree-sejour} (\textcolor{red}{d1} / \textcolor{blue}{duree} \\
                            \phantom{x}\hspace{3em}:\textcolor{green}{quantite} "cinq" \\
                            \phantom{x}\hspace{3em}:\textcolor{green}{unite} "jours") \\
                            \phantom{x}\hspace{2em}:\textcolor{green}{lieu} (\textcolor{red}{a1} / \textcolor{blue}{adresse} \\
                            \phantom{x}\hspace{3em}:\textcolor{green}{ville} "Paris"\\
                            \phantom{x}\hspace{3em}:\textcolor{green}{quartier}"huitième arrondissement")) \\
                            \phantom{x}\hspace{1em}:\textcolor{green}{etat} "en cours")
                            }
                \subcaption{Targeted semantic tree annotation.}
            \end{subfigure}
            
            \caption{Example of enriched annotation for a user turn of the MEDIA dataset. It can be translated as "I would like to book err two double bedrooms and one single bedroom for the five days of Christmas at Christmas so in the eighth district of Paris.". Node \textcolor{red}{identifiers} are in \textcolor{red}{red}, \textcolor{blue}{node types} in \textcolor{blue}{blue}, \textcolor{green}{relation types} in \textcolor{green}{green} and structure in black. Transcription spans are quoted.}
            \label{fig:old_to_new_MEDIA}
        \end{figure}

    % Recent initiatives towards spoken dialogue datasets
    % Difficulty to obtain rich (structured and contextual) annotations for spoken datasets
    Some recent datasets attempt to foster work on spoken dialogues: A spoken version of the MultiWOZ dataset \cite{Budzianowski2018MultiWOZA} with vocalized user turns was published in the context of the \textit{Speech Aware Dialogue Systems} track of the \nth{11} edition of the Dialogue System Technology Challenge\footnote{https://dstc11.dstc.community/} (DSTC11) \cite{soltau-etal-2023-dstc}. SpokenWOZ \cite{si2023spokenwoz} provides 5,700 dialogue recordings annotated with the same annotation format as MultiWOZ \textit{i.e.} list of slot-value pairs. However none are as fine-grained as Abstract Meaning Representation (AMR) \cite{banarescu2013abstract}. The STOP dataset \cite{tomasello2022stop} provides structured semantic trees for single turn voice commands which unfortunately leads to only 16\% truly requiring trees (\textit{i.e.} depth greater than 2) \cite{chen2020top}. Conversely, spoken TOD are well suited for structured semantic trees since sub-tasks are discussed in sub-dialogues which complement each other. 
    
    The high cost of fine-grained annotations seems to be the hurdle preventing spoken dialogue datasets from embracing such formalism. Yet, unannotated data is much more affordable and semi-supervised approaches gather a large panel of tasks~\cite{liao2019unsupervised,yang2022survey,georgila2009automatic}, leading to a low-cost compromise towards automatic annotation. 
    With the rise of large generative models, automatic annotation of textual data is thus reaching a new level of possibilities. Indeed, finetuning Large Language Models (LLM) on consumer-grade GPUs \cite{hu2022lora} is now a reality and has been applied to complex tasks such as Machine Translation \cite{moslem2023fine}. Moreover, prompt-based approaches \cite{savelka2023unlocking,zhang2024teleclass,gong2023lanser} even offer zero-shot annotations.
    Nevertheless, when dealing with complex tasks, fully automatic annotation remains challenging, because of behaviors such as hallucinations \cite{xu2024hallucination} and recent work often keeps a human in the loop \cite{kim2024meganno+} or go back to former semi-supervised methods.

\section{Method}

    Given the heavy cognitive load required to annotate dialogue turns with a contextual version of AMR, we attempt to automatize as much as we can while keeping high quality expectations. To do so we define a structured contextual meaning representation fitting the dataset's use case in section \ref{sec:ontology}, set an annotation pipeline in section \ref{sec:pipeline} and evaluate this system in section \ref{sec:evaluation}.

    \subsection{Structured Contextual Meaning Representation}
    \label{sec:ontology}
    
    Two adaptations are required to fit our dataset's use case: defining an ontology fitting the hotel booking task and handling cross-turn references. 
    
    The defined ontology comprises three type of concepts: domain related concepts (\textit{e.g.} hotel) which represent hotel reservation elements, operators (\textit{e.g.} et) which enable to apply an operation to other concept(s) and general purpose concepts (\textit{e.g.} adresse) which are domain-agnostic. Each concept has a unique identifier which enables exact cross-turn references. Concepts can be linked together or to transcription spans called literals. The edge's label depends on the pair of concepts considered. 
    
    Compared with the current labeled span annotation, this annotation takes into account the interactions between concepts and highlights the implicit concepts which are not uttered but group other concepts together (\textit{e.g.} hotel in Figure \ref{fig:old_to_new_MEDIA}). Those implicit concepts have higher error rates \cite{mdhaffar2022impact} and become essential to understand short turns with indirect references such as "the one with a swimming pool".

    \subsection{Annotation Pipeline}
    \label{sec:pipeline}
    % Annotation Pipeline
    % 1) Human annotations (annotation guide)
    % 2) LLM Finetuning (LoRA)
    % 3) LLM generation (grammar constrained, # max tokens)
    % 4) Annotation Evaluation (number of ontology errors and structural similarity)
    % Repeat

\begin{figure*}[htb!]
        \centering
        \includegraphics[width=\textwidth]{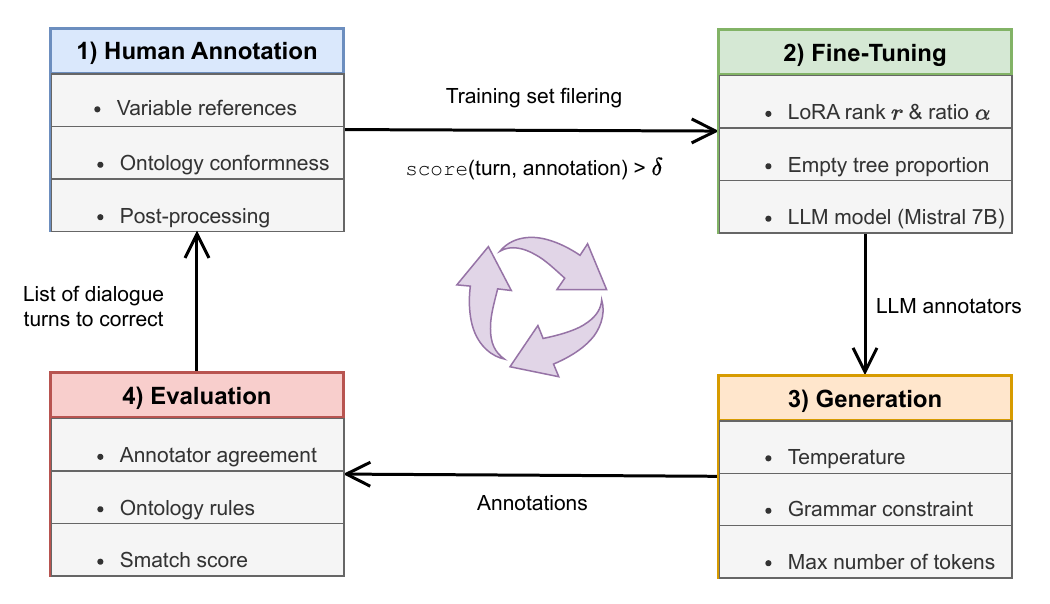}
        \caption{Overview of the semi-automatic annotation pipeline with the parameters of each step.}
        \label{fig:overview}
    \end{figure*}

    Our annotation pipeline comprises the following four steps which can be repeated over several iterations as illustrated in Figure \ref{fig:overview}.

    \subsubsection{Human annotation.}
    The first step consists in having 10 trained human annotators annotate a subset of the corpus following the defined ontology. We focused on the dataset's test set which comprises 208 dialogues with 30\% of them annotated by multiple annotators. The reported annotator agreement reaches an average semantic match (smatch) \cite{cai-knight-2013-smatch} of 77.28\%.  We present a few statistics of the obtained annotations in Table \ref{tab:annotation_stats} which highlight the importance of structured annotation. For each dialogue, we select the annotation with the least errors, according to a master-annotator, as the final one. Our goal is to leverage this set of clean annotations to produce the remaining annotations. 

    In order to track the quality of the automatic annotations, we further annotate another set of 22 dialogues to form a 10\% fold of unseen annotations.

    \begin{table}[htb!]
        \centering
        \begin{tabular}{c|c|c}
             \textbf{Avg. \# user turns} & \textbf{width $> 2$} & \textbf{depth $> 2$} \\ \midrule
             16.95 & 31.86\% & 24.34\% \\
        \end{tabular}
        \caption{Annotated dialogue turns statistics.}
        \label{tab:annotation_stats}
    \end{table}

    \subsubsection{Fine-Tuning.}
    We then fine-tune a Mistral-7B LLM \cite{jiang2023mistral} on the human annotated data with Low Rank Adaptation (LoRA) \cite{hu2022lora}. Given a pre-trained weight matrix $\mathbf{W} \in \mathbb{R}^{d \times k}$, we fine-tune a delta matrix $\mathbf{\Delta W} = \mathbf{B}\mathbf{A}$ such that $\mathbf{B} \in \mathbb{R}^{d \times r}$ and $\mathbf{A} \in \mathbb{R}^{r \times k}$ with $r \ll min(d, k)$. We then only train this factorized matrix $\mathbf{\Delta W}$ with the forward pass thus modified to:

    \begin{equation*}
        \mathbf{y} = \mathbf{W}\mathbf{x} + \frac{\alpha}{r} \mathbf{\Delta W}\mathbf{x}, \mbox{with } \alpha = mr, m \in \mathbb{N}
    \end{equation*}
    
    The LLM is fed the following prompt template in which it is tasked to provide a structured annotation of the last turn of a sequence of $t$ (\texttt{agent}, \texttt{user}) speaker turns transcription pairs\footnote{In practice the model is only fed the 5 previous turns to limit the number of tokens.}.

    \begin{tcolorbox}[colback=gray!30,%gray background
                  colframe=black!60,% black frame colour
                  width=\linewidth,
                  arc=3mm, auto outer arc,
                  title=Prompt Template
                 ]
        Below is an instruction that describes a task, paired with an input that provides further context. Write a response that appropriately completes the request. \\
        \#\#\# Instruction: Provide the tree annotation of what is said by the user in the given dialogue. \\
        \#\#\# Input: agent: [\texttt{agent}$_0$]; user: [\texttt{user}$_0$] \dots \ agent: [\texttt{agent}$_t$]; user: [\texttt{user}$_t$] \\ 
        \#\#\# Response:
    \end{tcolorbox}
    
    \subsubsection{Constrained Generation.}
    In order to reduce hallucinations and ensure that every output is correctly formatted while preserving originality, we implement a grammar constrained decoding \cite{geng-etal-2023-grammar} and allow up to 256 newly generated tokens. At each decoding step, the grammar constraint simply sets the log-probabilities of the forbidden tokens of the vocabulary to $-\infty$. The next token's probability distribution is thus restricted to grammar valid tokens. The main challenge consists in decoding smaller units (\textit{i.e.} tokens) than the grammar terminals while ensuring that the decoding remains in valid grammar states, especially when reaching literals which are supposed to be open vocabulary. In that case, our constraint decoding only allows tokens from the speakers turns until another quote is decoded. While more restrictive, this ensures that literals match speaker's transcription spans. Algorithm \ref{alg:constrained_decoding} presents how we select the set of allowed token level decoding paths to transition from one grammar valid state to another.

    \begin{algorithm}[htb!]
        \caption{Ensures that the decoding remains in valid grammar states.}
        \label{alg:constrained_decoding}
        \KwData{$W$ next grammar potential terminals, $T$ model's tokenizer, $eos$ end-of-speech token}
        \KwResult{$map$ allowed tokens mapping for all decoding paths which reach another valid grammar state}
        $map \gets $ empty map\;
        \eIf{$W$ is empty}{
            $map[T(eos)] \gets $ empty map\;
        }{\For{$w \in W$}{
            $tokens \gets T(w)$\;
            $next\_tokens \gets map$\;
            \While{$len(tokens) > 1$}{
                $t_0 \gets tokens[0]$\;
                \If{$t_0 \not\in next\_tokens$}{
                    $next\_tokens[t_0] \gets $ empty map\;
                }
                $tokens \gets tokens[1:]$\;
                $next\_tokens \gets next\_tokens[t_0]$\;
            }
            $next\_tokens[tokens[0]] \gets $ empty map\;
        }
        }
    \end{algorithm}

    \subsubsection{Annotation Evaluation.}
    Finally we evaluate the generated annotations with the AMR Semantic Match (smatch) score \cite{cai-knight-2013-smatch}. This metric searches for the best variable alignment and then computes the F1 score of the matching triples for this alignment. While it accounts for potential variable permutations, it remains a matching metric. Indeed, an error in the name of a concept or in a literal value invalidates the whole triple. Yet, averaged over many annotations, it provides a valuable insight in the structural quality of a set of annotations. It thus helps us to choose the best LLM annotator and to quantify the annotation quality over the held-out dialogues. Additionally, we track the number of ontology errors.  

\section{Results}

    \subsection{Dataset}

    The MEDIA dataset \cite{Devillers2004TheFM} comprises over 1,000 French hotel reservation dialogue recordings annotated with turn-level span concept labels over the manual transcriptions. It is recognized as a challenging spoken dialogue understanding dataset \cite{bechet2019benchmark}. Indeed the Wizard-of-Oz collection protocol provided loosely scripted scenarios which fostered rich and natural interactions. 

    \subsection{Experimental Setup}
    \label{sec:evaluation}
    % Is an LLM a "reliable" annotator: Did we transfer enough knwoledge from the clean set to the automatic set?
    % Training on the automatically annotated train set and comparing with the automatically annotated dev set and with the human annotated test set 

    This paper addresses three concerns around the annotation pipeline described in the sections above:
    \begin{enumerate}
        \item Is LLM fine-tuning relevant for this setup?
        \item Is the knowledge captured by the LLM effectively transferred to the automatic annotations?
        \item What does this mean for semi-automatic annotation?
    \end{enumerate}

    To assess the relevance of LLM fine-tuning, we compute pairwise annotation comparisons produced by a human A, a human B and a LLM X which is either fine-tuned or prompted in a few-shot manner.

    To evaluate the knowledge captured by the automatic annotations, we fine-tune the backbone LLM over the automatic annotations and compare its quality with the one which provided the annotations.

    Finally, to grasp the implications for semi-automatic annotation, we train a score estimator and use it to filter out the worst annotations (\textit{i.e.} score lower than the threshold $\delta$) before iterating over the model training and annotation. The estimator is composed of a Support Vector Regressor (SVR) over sentence embeddings\footnote{Obtained with the model at \href{https://huggingface.co/thenlper/gte-large}{https://huggingface.co/thenlper/gte-large}} of the last dialogue turn and the produced annotation.  

    \subsection{Relevance of LLM Fine-Tuning}
    We first compare a standard prompting strategy to fine-tuning one. Prompting can be done on any model including commercial ones which are often among the top performers of leaderboards. However, their token pricing policies may become prohibitive for iterative pipelines. On the other hand, fine-tuning requires a GPU and careful hyper-parameters setting but enables fine-grained customization.  
    
    \begin{figure}[htb!]
        \centering
        \includegraphics[width=\linewidth]{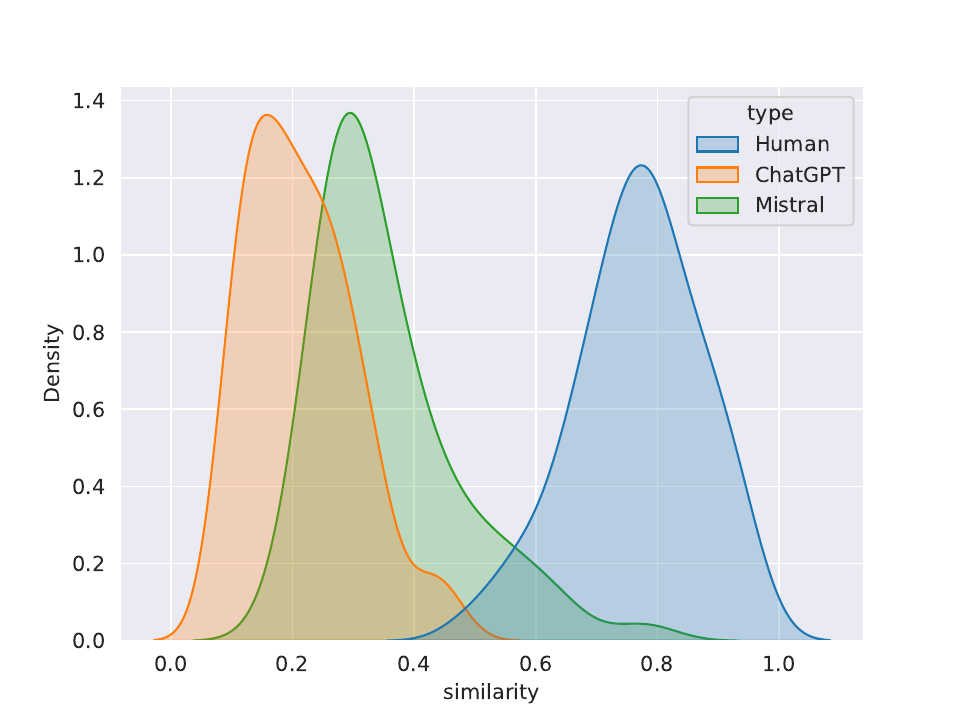}
        \caption{Smatch scores distribution of pairwise comparisons of human and automatic annotations.}
        \label{fig:similarities}
    \end{figure}
    
    Figure \ref{fig:similarities} presents the similarity (computed with smatch) distributions of pairwise comparisons between human and automatic annotations. We observe that a fine-tuned Mistral-7B can achieve higher similarities than Chat-GPT 3.5. However both LLMs remain far from human-wise annotation similarities indicating that iterative annotation should be promising. Therefore we choose the fine-tuning approach over the prompt-based one.

    \subsection{Knowledge captured by automatic annotations}

    We perform a hyper-parameter grid search over $\alpha \in [r, 2r]$, $r \in [16, 128, 512]$ and learning rate $\eta \in [1, 4, 8] \times 10^{-4}$ and select $r = \alpha = 512$ and $\eta = 4 \times 10^{-4}$. We present the results obtained with this model over the clean set and the unseen set in Table \ref{tab:knowledge_transfer}.

    \begin{table}[htb!]
        \centering
        \begin{tabular}{l|c|c|c|c}
        \textbf{Training set}  & \multicolumn{2}{c|}{\textbf{Clean}}  & \multicolumn{2}{c}{\textbf{Unseen 10\%}} \\ \midrule
        & \textbf{Full} & \textbf{Empty} & \textbf{Full} & \textbf{Empty} \\ \midrule
        \textbf{Clean last turn} & 52.43 \scriptsize{+/-18.82} & 94.78 \scriptsize{+/-9.94} & 33.12 \scriptsize{+/-12.57} & \textbf{90.06} \scriptsize{+/-11.73} \\
        \quad \textit{ w. grammar} & 73.31 \scriptsize{+/-10.84} & 0.0 & 60.94 \scriptsize{+/-10.02} & 0.0 \\
        \textbf{Clean history} & 86.85 \scriptsize{+/-10.06} & 92.99 \scriptsize{+/-11.31} & 58.44 \scriptsize{+/-13.29} & 78.2 \scriptsize{+/-23.83} \\
        \quad \textit{w. grammar} & 82.17 \scriptsize{+/-10.92} & 0.0 & \textbf{66.66} \scriptsize{+/-7.62} & 0.0 \\
        \textbf{Iteration 1} & 49.93 \scriptsize{+-15.93} & 86.56 \scriptsize{+-15.13} & 43.65 \scriptsize{+-14.66} & 81.69 \scriptsize{+-17.06} \\
        \quad \textit{w. grammar} & 64.36 \scriptsize{+-11.95} & 0.0 & 60.99 \scriptsize{+-13.0} & 0.0 \\
        \textbf{Merged} & 52.22 \scriptsize{+-15.09} & 86.56 \scriptsize{+-15.13} & 46.52 \scriptsize{+-14.31} & 81.69 \scriptsize{+-17.06} \\ \bottomrule   
        \end{tabular}
        \caption{Smatch scores for the same backbone model trained over different training sets and evaluated on the clean annotations and the unseen 10\% fold.}
        \label{tab:knowledge_transfer}
    \end{table}

    We first observe the importance of the dialogue history for such fine-grained annotations as the \textbf{Clean last turn} model which is only provided the last dialogue turn performs worse than the \textbf{Clean history} model. 
    
    Then, the grammar constrained decoding seems to behave as a complex temperature: it tends to make the model more verbose thus never predicting empty trees. This constrained decoding seems to disturb the calibration of the model's conditional probabilities. It improves the full tree annotations but makes the model more verbose hence reducing the performance on empty trees.

    Finally, the annotations produced by the \textbf{Clean history} model are of high enough quality to train well performing model. Indeed, when filtering out the worst examples as described above, the \textbf{Iteration 1} model comes close to its parent. Further iterations and more selective filtering therefore seem promising to improve the models' performances. In addition, the grammar constrained decoding seems more relevant with such noisier models. Merging both unconstrained and constrained predictions helps reach a happy-medium between full and empty trees as shown by the \textbf{Merged} model.

\section{Discussion}  

    This work focuses on the MEDIA dataset \cite{Devillers2004TheFM} since it is recognized as challenging \cite{bechet2019benchmark} and hence provides complex enough situations to require fine-grained annotations. However it is a single domain French dataset. Performances might be better on higher resources languages such as English and worse on lower resources languages. While this paper focuses on ontology specific to MEDIA's use case, designing a similar one for multi-domain use-cases should be manageable in most cases.

    With the recent hype around Large Language Models, a huge quantity of resources has been made available. This paper only focuses on well-established models and techniques. To the best of our knowledge prompt design remains an unsettled topic. We thus experimented with several formulations and selected the one which seemed the most promising. There might be better formulations than the one proposed in this paper. 

\section{Conclusion}

    Our journey towards fine-grained annotations for spoken dialogue datasets has led us to the realm of LLMs in order to comply with our low annotation budget. This exploratory work provides valuable insights to the community on the design of complex automatic annotation for spoken dialogue datasets. Indeed, our method may accelerate manual annotation and/or be included in a fully automated setting by carefully selecting the training examples. 
    We highlight that open-weights LLMs fine-tuning is relevant for this annotation setup since it enables faster iterations than prompting commercial models while remaining competitive. We also propose a grammar constrained decoding strategy which struggles with non-informative dialogue turns but improves the annotation of correctly annotated dialogue turns. Our produced annotations contain part of the knowledge from the human annotations, and can be effectively learned by a model.

\begin{credits}

\subsubsection{\discintname}
The authors have no competing interests to declare that are relevant to the content of this article.
\end{credits}
\bibliographystyle{splncs04}
\bibliography{mybib}
\end{document}